# Enhancing Deep Neural Network Training Efficiency and Performance through Linear Prediction

Hejie Ying[1,2], Mengmeng Song[1,2]*, Yaohong Tang[1,2],
Shungen Xiao[1,2], Zimin Xiao[1],
1 Ningde Normal University, No. 1 College Road, Ningde, 352101, FuJian, China.
2 New Energy Vehicle Motor Industry Technology Development Base,
Ningde Normal University, No. 1 College Road, Ningde, 352101, FuJian, China.
*Corresponding author.
E-mail(s): t2135@ndnu.edu.cn

**Abstract:**

   Deep neural networks have achieved remarkable success in various fields. However, training an effective deep neural network still poses challenges. This paper aims to propose a method to optimize the training effectiveness of deep neural networks, with the goal of improving their performance. Firstly, based on the observation that parameters (weights and bias) of deep neural network change in certain rules during training process, the potential of parameters prediction for improving training efficiency is discovered. Secondly, the potential of parameters prediction to improve the performance of deep neural network by noise injection introduced by prediction errors is revealed. And then, considering the limitations comprehensively, a deep neural network Parameters Linear Prediction method is exploit. Finally, performance and hyperparameter sensitivity validations are carried out on some representative backbones. Experimental results show that by employing proposed Parameters Linear Prediction method, as opposed to SGD, has led to an approximate 1% increase in accuracy for optimal model, along with a reduction of about 0.01 in top-1/top-5 error. Moreover, it also exhibits stable performance under various hyperparameter settings, shown the effectiveness of the proposed method and validated its capacity in enhancing network's training efficiency and performance.

# 1. Introduction

From epoch-making Convolutional Neural Network (CNN) [1] to Deep Belief Network (DBN) [2] and various effective and remarkable neural network structures [3],[4],[5],[6],[7],[8],[9], Deep-learning Neural Networks (DNN) today has undoubtedly become the mainstream of Machine Learnings, and have demonstrated remarkable success in tasks such as computer vision, natural language processing, and speech recognition. However, as DNN models grow larger and more complex, training DNN models remains a challenging and time-consuming task, often requiring extensive computational resources and careful hyperparameter tuning.

The training effectiveness of DNN models is crucial for their success and widespread adoption. Despite their impressive capabilities, DNN are susceptible to several challenges that can hinder their training and limit their performance. One of the main concerns among these challenges is parameters optimization problem. Lots of remarkable works have been proposed to optimizing this process based on Gradient Descent, among which including the non-adaptive method from SGD to DEMON [10], [11], and the adaptive method from AdaGrad to AdamW [12], [13], [14], [15], [16].

However, even equipped with the methods above, researchers still have to carefully tuning hyperparameters to struggle for the 1% accuracy improvement, because the better performance of DNN usually comes at the availability of exceptionally large computational resources on specialized hardware accelerators in model training process, which necessitate similarly substantial energy consumption [17], [18]. Moreover, the extensive periods need for DNN model training also prolong the validation cycle of the algorithm, which often result in researchers invest time and effort in vain. Therefore, research on how to improve the training efficiency and performance of DNN is still necessary.

To get better DNN training efficiency and performance, in this paper, focusing on the parameters optimization problems, proposed a DNN Parameters (weights and bias) Linear Prediction (PLP) method, aims to predict them according to their tendency during training directly, but not just using SGD to find the optimal parameters step by step, or introducing new momentum or learning rate updating strategies to get them in-directly [10], [11], [12], [13], [14], [15], [16]. Specifically, proposed PLP method takes every 3 iterations a cycle. Firstly, stores the first 3 iteration results of parameters get from SGD process. Secondly, calculate the slope of the median line of the triangle formed by the first 3 iteration results, and take the midpoint of the last 2 stored results as the start point. And then make the linear prediction for the parameters using the slope and the start point. Finally update the predict parameters to the model for the next optimization iteration.

With the performance and sensitivity evaluation experiments, the proposed PLP method is proved to be hyperparameters insensitive, and be able to get better training performance compare to SGD (lower average loss and higher accuracy on validation set, higher accuracy and lower top-1/top-5 error on test set), and close or even better results compare to DEMON method which is the SOTA non-adaptive method. Also, with these experiments, the proposed method is proved to be general and can be extended to other tasks with minor effort.

The rest of this paper is organized as follows. In section 2, relate works that aim to improve DNN training efficiency and performance are introduced. In section 3, we detailed the implementations of proposed PLP method, and compare it with existing methods in terms of complexity and memory requirements. In section 4, performance and sensitivity evaluation experiments results on some representative backbones (Vgg, Resnet and GoogLeNet) are provided and discussed. Finally, conclusions are drawn in section 5.

## 2. Related Works

Researchers have proposed various methods in order to optimize the parameters optimization problem for better DNN training efficiency and performance. Those methods can be roughly divided into 2 types: non-adaptive and adaptive.

The representative method of non-adaptive method is SGD, from which a variety of improved algorithms have been derived, DEMON which is a momentum decay rule that reduces the contribution of gradients to future updates, while incurring minimal computational overhead is the latest and widely used one among them [11]. Additionally, Tran T et.al [19] have introduced a Shuffling Momentum Gradient method that combines shuffling strategies and momentum techniques to accelerate the optimization process. By using difference between the adjacent mini-batch gradient to update the exponential moving averages of the gradient variation, Wei Y et.al [20] have proposed a stochastic gradient descent with momentum and difference. On the basis of the observation that the convergence rate's upper bound is relate to current learning rate, Wang K et.al [21] have summarized the expression of the current learning rate determined by historical learning rates, and applied it to SGD method showing better performance. Research [22] used momentum gradient descent in the early stages of training and switches to plain gradient descent in the later stages with scaling, experiments shown this method has faster training speed, higher accuracy and better stability.

Adaptive methods represented by Adam have received increasing attention these years due to its good convergence performance and robustness. AdaGrad [12], AdaDelta [13] and Adam [18] are most widely used. To combine the benefits of both adaptive learning rate methods and momentum-based methods leads to Adam [18]. QHAdam extends the Adam algorithm by introducing a quasi-hyperbolic momentum term which aims to strike a balance between the benefits of adaptive and the stability achieved by non-adaptive methods [15]. By incorporating weight decay directly into the parameters update step, AdamW helps prevent overfitting by adding a penalty term to the loss function based on the magnitudes of the model's weights [16]. Besides these mainstream works, other studies have also been widely conducted. Liu H et.al [23] have combined momentum with adaptive gradient descent which incorporates a running sum of the transformed gradient with element-wise adaptive gradient descent, and resulted in faster convergence. Chen J et.al [24] have proposed a consensus-based global optimization method with adaptive momentum estimation that can handle non-differentiable activation functions and thus approximate low regularity functions with better accuracy. In research [25], to overcome the overshoot phenomenon of Adam, proportional P and derivative D of PID controller were borrowed, and the Adam was derived into the integral I component, this adaptive-PID method have shown better convergence rate and accuracy compare to Adam. Some works also attempted to adopt different strategies at different stages to take advantages of both methods [26].

The adaptive methods are able to get faster convergence speed but relatively lower convergence accuracy. Conversely, the non-adaptive methods usually convergence slower, but with higher convergence accuracy [20], [26]. Additionally, both the adaptive and non-adaptive methods now prefer to introducing new momentum and learning rate adjust strategies or algorithms to achieve better training results, these have inevitably increased system complexity while weakens the interpretability of the parameters optimization process, especially adaptive methods [27]. Issues above have explained why the non-adaptive optimizers are usually the default setting in most DNNs, especially SGD.

Hence, the exploration of relatively simple but effective ways to enhance DNN training efficiency and performance is still meaningful.

# 3. Proposed Method

In this section, we first discuss the reason why proposed PLP method works. And then the implementation details of proposed method are introduced. After that, we compare it with existing methods in terms of complexity and memory requirements.

## 3.1 Problem Formulation

As shown in Fig.1, it is the changing curve of a weight and bias that are randomly picked from the training process of a Vgg16 network trained on CIFAR-100 dataset. It's obvious that although the changes of parameters exhibit some instability due to the adoption of SGD, but from the smoothed data curve, it is not difficult to learn that their change still shows a certain regularity, which show the feasibility of parameters prediction to improve the training performance of DNN. In other words, if we can predict the value for each parameter during training process accordingly, then it is possible to achieve better training efficiency.

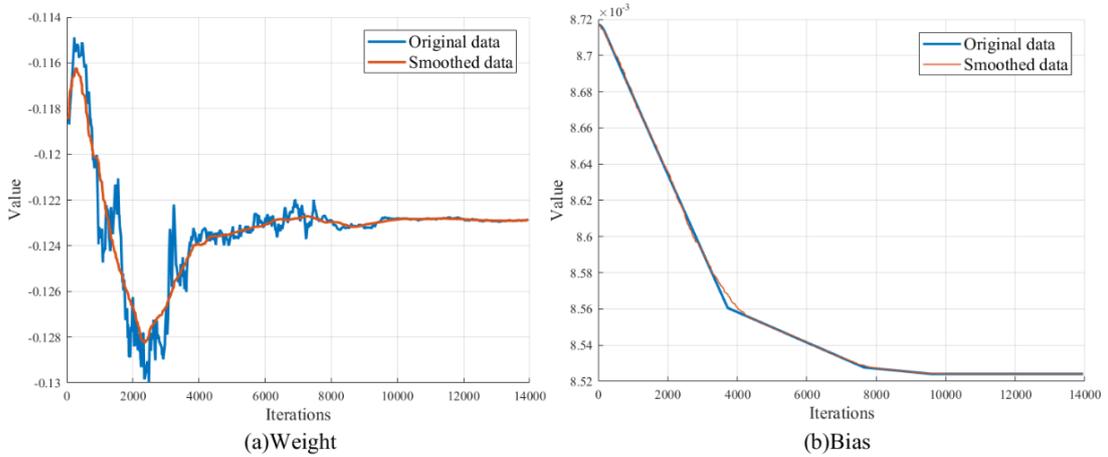

Figure 1: Example of parameters changing tendency during Vgg16 training.

Meanwhile, due to the characteristics of SGD or different settings of DNN, such as parameters initial values, learning rate or batch-size, the same parameters in the same DNN train with the same dataset may show different changing rules. Based on this, considering the growing number of DNN parameters nowadays, training specific prediction models for each parameter in real-time during the training process with more accurate regression algorithms like Least Squares or Support Vector Regressions is not possible. Additionally, since SGD inherently allows gradient computations with noise [28], [29], the SGD process itself has certain tolerance for the loss of parameters prediction accuracy. Consequently, the real-time performance and computational complexity are the first concerns of parameters prediction methods, but there can be certain trade-offs in accuracy.

Furthermore, considering generalization effect of noise injection to DNN model training [30],[31], the introduction of noise caused by parameters prediction error has the effect on model generalization, and also have the potential to help the SGD optimization process jump out of the local optimal solution. Therefore, it is trustworthy that parameters prediction in DNN models can also be able to improve the performance of DNN training model in the aspect of accuracy and generalization capability.

Based on the discussions above, we proposed a Parameters Linear Prediction (PLP) method for DNN parameters prediction.

### 3.2 Parameters Linear Prediction

The pseudocode of the PLP method is given in Algorithm 1.

| Algorithm 1 PLP method |
| --- |
| **Require:** SGD with momentum 0.9. $T$ : total number of iterations |
| **for** $t = 1$ to $T$ **do** |
|     Model training with SGD                                                   0. Model training |
|    **if** iteration % 3 != 0 **then** |
|       storage[iteration%3] $\leftarrow$ parameters            1. Parameters storage |
|    **else then** |
|       $m_{12} = \dfrac{storage[0] + storage[1]}{2}$,   $m_{23} = \dfrac{storage[1] + storage[2]}{2}$ |
|       slope = $m_{23} - m_{12}$                                2. Slope & start point calculation |
|       Predicted_param = $m_{23}$ + slope*step          3. Linear prediction |
|       Model $\leftarrow$ Predicted_param                      4. Parameters updating |
| **end for** |

As show in Algorithm 1. PLP method takes every 3 iterations a cycle (iteration refers to a mini-batch go through a complete forward and backward propagation), and the whole process starts from step "0. model training" to step "4. Parameters updating".

Specifically, in each cycle, firstly, the PLP method loop between "0. model training" and "1. Parameters storage" until it stores the first 3 iteration results of parameters $w$n_1, $b$n_1, $w$n_2, $b$n_2 and $w$n_3, $b$n_3 calculate by SGD, here n refers to *n*-th layer of model and the number after it represents the value stored in the *i*-th iteration.

Secondly, in "2. Slope & start point calculation", the midpoint $m_{12}$ and $m_{23}$ between two pairs of results stored in "1. Parameters storage" are calculated as follows (take weights as example, the calculation for bias is exactly the same):

$$m_{12} = \frac{wn\_1 + wn\_2}{2} \tag{1}$$

$$m_{23} = \frac{wn\_2 + wn\_3}{2} \tag{2}$$

And then, as shown in eq.3, calculate the slope of median line of the triangle formed by the first 3 results using the midpoint $m_{12}$ and $m_{23}$, because the interval between $m_{12}$ and $m_{23}$ here is equivalent to 1 iteration, the denominator of slope is omitted.

$$\text{slope} = m_{23} - m_{12} \tag{3}$$

Afterward, in "3. Linear prediction", taking the midpoint $m_{23}$ as the starting point for linear prediction and make the linear prediction for the parameters using the slops and the start point as shown in eq.4 (take weights as example, the prediction for bias is exactly the same). The "step" in eq.4 refers to the number of steps to predict, to avoid introducing too much prediction error which may lead to convergence problems, "step" is set to 1 in this paper.

$$wn\_4 = m_{23} + step * slop \tag{3}$$

Finally, in "4. Parameters updating", the predicted parameters are updated to the model for the next optimization iteration.

For clarity, the overall process is also shown in Fig.2.

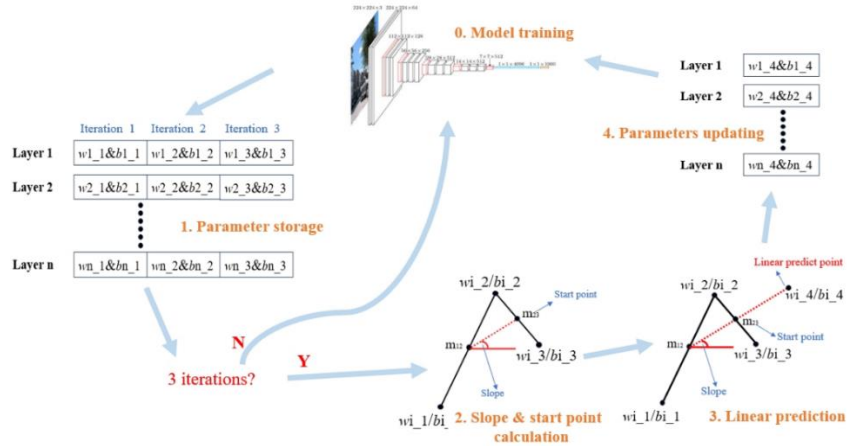

Figure 2: Overview of Linear Prediction method.

What needs to be added is that the proposed method takes 3 results for prediction instead of 2 or more because the 3-point prediction is smoother than the 2-point prediction, so that the prediction errors are smaller, and also saves more memory and computational power than the method using more points.

With the introductions above, it's obvious that, different from the mainstream parameters optimization methods mentioned in Section 2 that prefer to introducing new methods or algorithms for momentum or learning rate adjustment, the proposed method focuses on the parameters themselves, attempting to accelerate model convergence through linear prediction, which is relatively simple and interpretable. Specifically, algorithms below are the representative methods of adaptive and non-adaptive parameter optimization methods: Adam and DEMON [14],[11].

| **Algorithm** Adam | **Algorithm** DEMON |
|---|---|
| **Require:** … | **Require:** … |
| while $\theta_t$ not converged **do** <br><br> $m_t = \beta_1 \cdot m_{t-1} + (1-\beta_2) \cdot g_t$ <br> $v_t = \beta_2 \cdot v_{t-1} + (1-\beta_2) \cdot g_t^2$ <br> $\overline{m_t} = m_t / (1-\beta_1^t)$ <br> $\overline{v_t} = v_t / (1-\beta_2^t)$ <br> $\theta_t = \theta_{t-1} - \alpha \cdot \overline{m_t} / \left(\sqrt{\overline{v_t}} + \varepsilon\right)$ <br><br> **end while** | **for** $t = 0, \ldots, T$ **do** <br><br> $\beta_t = \beta_{init} \cdot \dfrac{(1-t/T)}{(1-\beta_{init}) + \beta_{init} \cdot (1-t/T)}$ <br> $\theta_{t+1} = \theta_t - \eta \cdot g_t + \beta_t \cdot v_t$ <br> $v_{t+1} = \beta_t \cdot v_t - \eta \cdot g_t$ <br><br> **end for** |

It can be inferred that the proposed PLP method has time complexity $O(N)$ and spatial complexity $O(N)$ (N refers to the number of DNN parameters), which is the same as DEMON and Adam methods. However, it is obvious that the computational complexity of PLP method is relatively lower as it has fewer arithmetic operations. And during calculation process, the memory requirement for the intermediate variables of PLP method is 7*N (*storage*[3], $m_{12}$, $m_{23}$, *slope* and *Predicted_param*), which can be optimized to 4*N (*tmp$_1$*, *tmp$_2$*, *tmp$_3$* and *Predicted_param*), while that of Adam is 6*N ($g_t$, $m_t$, $v_t$, $\overline{m_t}$, $\overline{v_t}$ and $\theta_t$), Demon is 3*N ($g_t$, $v_t$ and $\theta_t$). The memory requirement of PLP method is only slightly larger than that of DEMON and smaller than Adam.

In general, compared with the mainstream methods, PLP has the advantage of computational complexity, but only slight disadvantages in memory requirements. Moreover, proposed PLP method only performs once every 3 iterations instead of each iteration, so that effectively reducing the computational load compared to other methods.

## 4. Experiments

In this section, experimental details are described. And then we perform experiments to evaluate the performance and hyperparameter sensitivity of proposed method.

### 4.1 Implementation Details

We experimented with 3 representative backbones (i.e., Vgg16 [4], Resnet18 [6] and GoogLeNet [5]) and 2 representative non-adaptive parameters optimization methods (i.e., SGD and DEMON). All networks are randomly initialized under default settings of PyTorch with no pretraining on any external dataset. Methods are evaluated on these networks with CIFAR-100 dataset, which has 100 categories, each containing 600 images, there are 50000 training images and 10000 test images. 20% of the training set has been randomly split into validation set. And a NVIDIA GeForce RTX 3060 Laptop GPU is used to implement and evaluate the experiments.

The codes were implemented based on Python 3.10.9 with Pytorch 2.0.1. All models were trained 100 epochs with different parameters optimization methods. The weight decay is 1e-4 and momentum is set to 0.9. As for learning rate, in performance evaluation, cyclic learning rate update strategy with base learning rate of 0.001 and max learning rate 0.002 was adopted to achieve more stable results. And no learning rate adjustment strategy in sensitivity evaluation.

To make the validation of the PLP method more objective and avoid erroneous validation results caused by randomness, the accuracy and top-1/top-5 error evaluations in this section were repeated 10 times and we reported the average of them (i.e., we trained 10 models for each of the 3 selected networks using the proposed PLP method, SGD and DEMON respectively, and tested each them on the test set).

### 4.2 Performance Evaluation & Analyzation

Fig.3 shows the accuracy of proposed PLP method, DEMON and SGD on training set. It can be seen that the accuracy of proposed PLP method and DEMON are almost overlapping, and both are superior to SGD in almost the whole training process. Combined with the loss curve on validation set shown in Fig.4, it is not difficult to draw a conclusion that the model is about to overfit around 40th epoch, and this has explained why the difference in training accuracy among 3 methods getting smaller and smaller after 40th epoch, that is, because the parameters in the overfitting stage gradually converge, leading to the performance of SGD gradually equalize to PLP and DEMON.

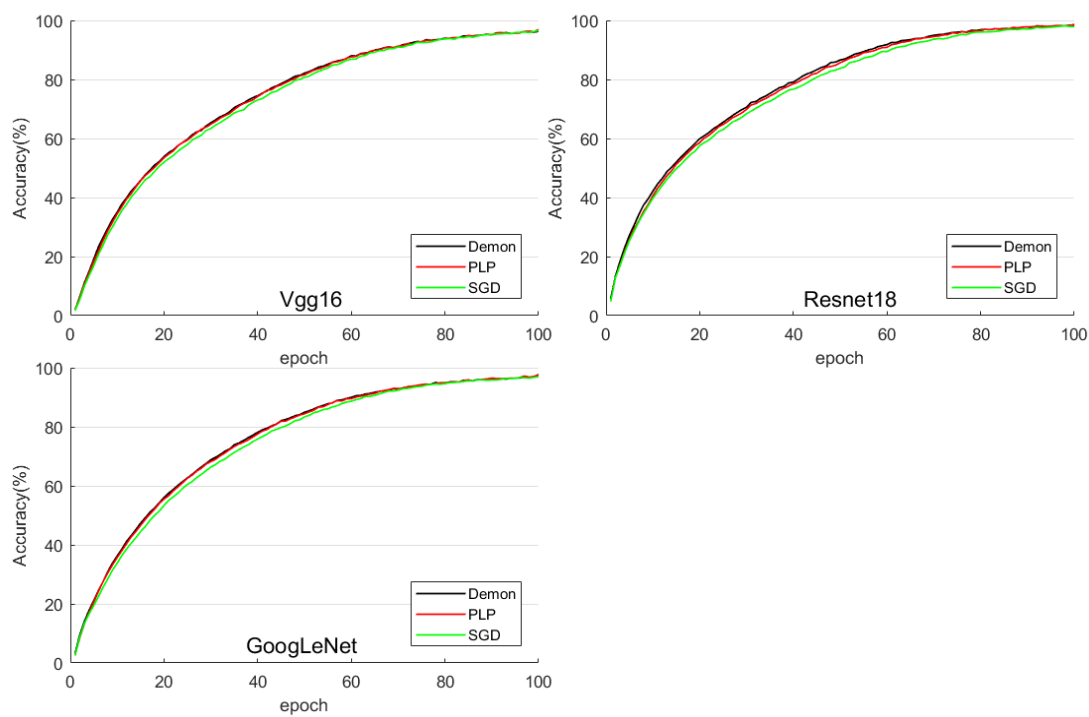

Figure 3: Accuracy and comparison of proposed PLP method, DEMON and SGD.

Fig.4 shows the loss and comparison of proposed PLP method, DEMON and SGD on validation set. As can be seen from Fig.4-"Loss Diff"-"SGD-PLP", there are some outliers before 40th epoch, where the proposed PLP method showing worse performance compare to SGD, i.e., the validation loss of proposed PLP method is larger than model with SGD. The root cause of these outliers are the characteristics of SGD that may lead to sudden changes in parameters values, result in proposed PLP method unable to follow its changes. Besides this, except for the overfitting stage after about 40th epoch, proposed method shows better performance during model training compared to SGD. Fig.4-"Loss Diff"-"PLP-DEMON" shown the loss comparison between PLP and DEMON, in the beginning of the training, DEMON method got lower loss, which are turned the other way around after about 20th epoch, showcased that the PLP method has better optimization performance compared to DEMON as parameters optimization deepens.

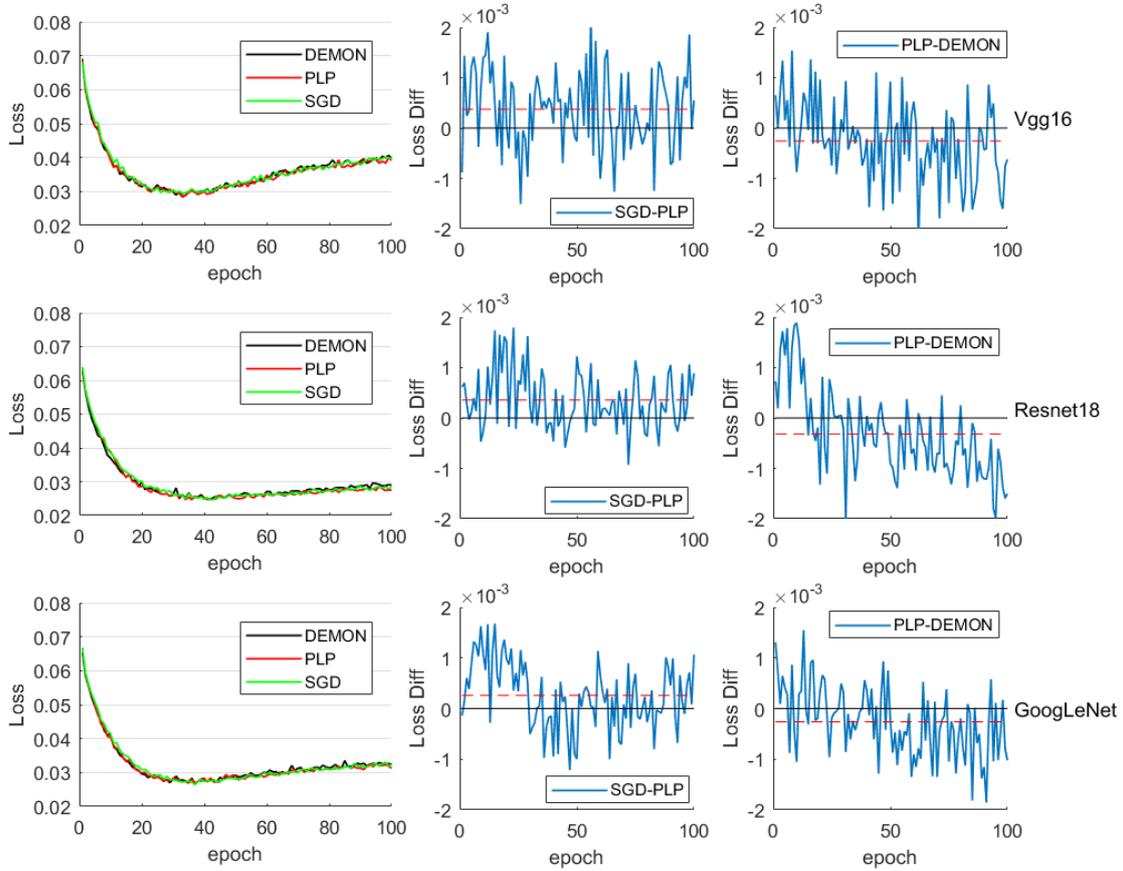

Figure 4: Loss and comparison on validation set of proposed PLP method, DEMON and SGD.

Fig.3 and Fig.4 have shown the ability of proposed PLP method in improving DNN training performance. To better illustrate the capacity of proposed method in improving the training efficiency of DNNs, Table 1, Table 2 and Table 3 have shown the epoch and loss values corresponding to the optimal training model obtained from 10 times tests for the 3 selected models.

Table 1 shows the comparison between PLP, DEMON and SGD based on Vgg16 net. Except in 1st and 8th tests the PLP method obtain higher loss in later epochs, and in 10th test lower loss is obtained in later epochs. In the rest of tests, the PLP method is able to get better performance in earlier epochs compare to SGD. The reason for the abnormality of the 1st, 8th and 10th tests is that the randomness of SGD may lead to the gradient mutation of parameters during training process, resulting in the noise introduced by PLP method exceeding the tolerance of SGD, which makes PLP method spend longer training epochs than normal method to converge. However, even under this condition, due to the introduction of noise, PLP may still be able to get better training performance with the generalization effect caused by the introduction of noise (10th test). As for comparison with DEMON, about 50% of the tests (1st, 5th, 7th, 8th, 10th) PLP converge faster with lower loss than DEMON, and converge slower with lower loss in 6th test, and in the rest 40% tests PLP performed worse. Considering the impact of randomness in the SGD process, based on the experimental results, it can be concluded that the proposed PLP method has demonstrated comparable performance to DEMON substantially.

Table.1 Vgg16 based comparison (factor of loss: 1e-2). Red indicates Top-1 performance

|  |  | 1 | | 2 | | 3 | | 4 | | 5 | |
|---|---|---|---|---|---|---|---|---|---|---|---|
|  |  | epoch | loss | epoch | loss | epoch | loss | epoch | loss | epoch | loss |
| Vgg16 | DEMON | 38 | 2.95 | 32 | 2.738 | 30 | 2.851 | 31 | 2.841 | 35 | 2.909 |
|  | PLP | 35 | 2.916 | 31 | 2.762 | 34 | 2.86 | 29 | 2.857 | 31 | 2.907 |
|  | SGD | 35 | 2.857 | 33 | 2.846 | 35 | 2.942 | 31 | 2.89 | 34 | 2.953 |
|  |  | 6 | | 7 | | 8 | | 9 | | 10 | |
|  | DEMON | 29 | 2.927 | 33 | 2.843 | 34 | 2.907 | 34 | 2.857 | 35 | 2.914 |
|  | PLP | 32 | 2.896 | 30 | 2.829 | 34 | 2.984 | 33 | 2.889 | 34 | 2.801 |
|  | SGD | 35 | 2.913 | 34 | 2.887 | 33 | 2.944 | 35 | 2.907 | 37 | 2.957 |

Table 2 and Table 3 show the comparison between PLP, DEMON and SGD based on Resnet18 and GoogLeNet. As shown in the tables, there are also some outliers like Table1 mentioned above, but overall, in most cases, the PLP method can obtain the optimal model faster than SGD during training process, and have similar performance to DEMON, verified the capacity of proposed method in improving the training efficiency of DNNs.

Table.2 Resnet18 based comparison (factor of loss: 1e-2). Red indicates Top-1 performance

|  |  | 1 | | 2 | | 3 | | 4 | | 5 | |
|---|---|---|---|---|---|---|---|---|---|---|---|
|  |  | epoch | loss | epoch | loss | epoch | loss | epoch | loss | epoch | loss |
| ResNet18 | DEMON | 39 | 2.518 | 40 | 2.487 | 36 | 2.516 | 37 | 2.511 | 38 | 2.521 |
|  | PLP | 42 | 2.492 | 39 | 2.455 | 38 | 2.474 | 39 | 2.525 | 36 | 2.483 |
|  | SGD | 45 | 2.502 | 35 | 2.485 | 38 | 2.520 | 37 | 2.537 | 40 | 2.533 |
|  |  | 6 | | 7 | | 8 | | 9 | | 10 | |
|  | DEMON | 40 | 2.486 | 42 | 2.543 | 38 | 2.513 | 37 | 2.522 | 39 | 2.496 |
|  | PLP | 39 | 2.499 | 44 | 2.515 | 40 | 2.461 | 35 | 2.517 | 38 | 2.487 |
|  | SGD | 41 | 2.527 | 48 | 2.548 | 45 | 2.529 | 36 | 2.534 | 42 | 2.507 |

Table.3 GoogLeNet based comparison (factor of loss: 1e-2). Red indicates Top-1 performance

| | | 1 | | 2 | | 3 | | 4 | | 5 | |
|---|---|---|---|---|---|---|---|---|---|---|---|
| | | epoch | loss | epoch | loss | epoch | loss | epoch | loss | epoch | loss |
| GoogLeNet | DEMON | 37 | 2.69 | 38 | 2.684 | 35 | 2.651 | 38 | 2.79 | 34 | 2.617 |
| | PLP | 37 | 2.684 | 37 | 2.677 | 35 | 2.588 | 41 | 2.842 | 38 | 2.667 |
| | SGD | 38 | 2.723 | 40 | 2.711 | 37 | 2.683 | 39 | 2.839 | 38 | 2.678 |
| | | 6 | | 7 | | 8 | | 9 | | 10 | |
| | DEMON | 36 | 2.641 | 38 | 2.702 | 37 | 2.694 | 34 | 2.626 | 38 | 2.683 |
| | PLP | 37 | 2.653 | 35 | 2.689 | 42 | 2.735 | 33 | 2.637 | 39 | 2.674 |
| | SGD | 39 | 2.71 | 35 | 2.751 | 39 | 2.718 | 38 | 2.702 | 39 | 2.723 |

To further demonstrate the effectiveness of the proposed method in enhancing DNN training efficiency and performance, with CIFAR-100 test set, the average accuracy, top-1/top-5 error of optimal models and the accuracy of model on constant training duration are shown in Table 4. As can be seen, the model trained with proposed PLP method had obtained higher accuracy than SGD at each stage of model training, with more than 1% accuracy improvement in average, showing that the better training performance of proposed PLP method is not achieved at the expense of generalization, but even achieving better generalization than baseline models. Also, the proposed PLP method get smaller top-1/top-5 error in average, further proved the effectiveness of proposed PLP method in getting better training performance.

And Compared to DEMON, both the accuracy on constant training duration and the accuracy and top1/5 error of optimal models are quite close, proved that PLP method and DEMON are well-matched in performance and efficiency, and both are superior to SGD.

Table 4. Average accuracy and top-1/top-5 error results on test set. Red indicates Top-1 performance

| Model | Epochs (Accuracy) | | | Optimal Model | | |
|---|---|---|---|---|---|---|
| | 10 | 20 | 30 | Accuracy | Top-1 error | Top-5 error |
| Vgg16(DEMON) | 37.91% | 50.34% | 56.83% | 57.38% | 0.4447 | 0.1632 |
| Vgg16(PLP) | 37.88% | 50.80% | 56.79% | 57.31% | 0.4317 | 0.1617 |
| Vgg16(SGD) | 36.15% | 49.33% | 55.51% | 56.42% | 0.4511 | 0.1654 |
| Resnet18(DEMON) | 38.01% | 50.97% | 57.10% | 59.88% | 0.4290 | 0.1563 |
| Resnet18(PLP) | 37.82% | 50.75% | 56.98% | 60.01% | 0.4312 | 0.1603 |
| Resnet18(SGD) | 35.62% | 49.86% | 56.16% | 58.76% | 0.4384 | 0.1640 |

| | | | | | |
|---|---|---|---|---|---|
| GoogLeNet(DEMON) | 37.31% | 51.45% | 56.13% | 59.86% | 0.4387 | 0.1642 |
| GoogLeNet(PLP) | 37.51% | 51.03% | 56.67% | 58.37% | 0.4333 | 0.1606 |
| GoogLeNet(SGD) | 34.95% | 49.65% | 55.79% | 57.70% | 0.4421 | 0.1664 |

Experimental results above suggest that the proposed PLP method is superior to SGD and comparable to SOTA non-adaptive method (DEMON) in terms of convergence speed and accuracy in most cases, shown the ability of proposed method in enhancing DNN training efficiency and performance. The existence of outliers also revealed a certain dependence of proposed PLP method on the reliability of predictions.

## 4.3 Sensitivity Evaluation & Analyzation

To demonstrate the hyperparameter sensitivity of proposed PLP method, we evaluated and analyzed the performance of PLP method with different learning rates and batch sizes on different backbones.

As introduced in Section 3.1, the regular variation of DNN parameters during the optimization process is the basis of the proposed PLP method. This poses certain requirements for the stability of parameter changes during the DNN optimization process when applying the PLP method. It is not difficult to conclude that if the learning rate is large and causes oscillations during the parameter optimization process, linear parameter prediction will introduce larger prediction errors, thereby weakening the performance of the PLP method. This is consistent with the experimental results shown in Table 5, which shown that the PLP method usually performs better with smaller learning rates (<0.01). To summarize, despite of the GoogLenet with learning rate 0.01, the proposed PLP method still shows better performance in DNN training efficiency or effectiveness in other cases, proving its stability for different learning rates.

Table 5. Performance comparison with different learning rate. Red indicates Top-1 performance

| LR | | VGG16 | | | Resnet18 | | | GoogLenet | | |
|---|---|---|---|---|---|---|---|---|---|---|
| | | DEMON | PLP | SGD | DEMON | PLP | SGD | DEMON | PLP | SGD |
| 0.01 | epoch | 40 | 44 | 41 | 26 | 25 | 27 | 38 | 41 | 45 |
| | Acc/% | 62.88 | 63.25 | 62.50 | 63.67 | 63.50 | 63.46 | 66.29 | 66.15 | 67.05 |
| | top-1 | 0.3812 | 0.3675 | 0.3750 | 0.3633 | 0.3657 | 0.3654 | 0.3371 | 0.3385 | 0.3259 |
| | top-5 | 0.1295 | 0.1218 | 0.1249 | 0.1178 | 0.1179 | 0.1183 | 0.1035 | 0.1044 | 0.1054 |
| 0.001 | epoch | 34 | 35 | 37 | 46 | 46 | 48 | 40 | 40 | 44 |
|        | Acc/% | 56.92  | 57.10  | 56.16  | 61.24  | 61.67  | 60.21  | 58.57  | 59.54  | 58.38  |
|        | top-1 | 0.4308 | 0.4290 | 0.4384 | 0.3976 | 0.3833 | 0.3979 | 0.4143 | 0.4046 | 0.4162 |
|        | top-5 | 0.1602 | 0.1600 | 0.1657 | 0.1326 | 0.1326 | 0.1316 | 0.1518 | 0.1416 | 0.1554 |
|        | epoch | 98     | 96     | 103    | 183    | 181    | 186    | 168    | 162    | 174    |
| 0.0001 | Acc/% | 48.15  | 49.22  | 49.19  | 55.17  | 55.73  | 55.02  | 49.76  | 50.99  | 49.57  |
|        | top-1 | 0.5185 | 0.5078 | 0.5081 | 0.4483 | 0.4427 | 0.4491 | 0.5024 | 0.4901 | 0.5043 |
|        | top-5 | 0.2276 | 0.2210 | 0.2199 | 0.1744 | 0.1680 | 0.1735 | 0.2176 | 0.2161 | 0.2203 |

Table 6 shown the comparison with some commonly used batch size values. Experimental results shown that the proposed PLP method exhibits good training performance with different batch size settings on different backbones, indicating the stability of the proposed method for the hyperparameter batch size.

Table 6. Performance comparison with different batch size. Red indicates Top-1 performance

|            |       | VGG16  |        |        | Resnet18 |        |        | GoogLenet |        |        |
|------------|-------|--------|--------|--------|----------|--------|--------|-----------|--------|--------|
| Batch size |       | DEMON  | PLP    | SGD    | DEMON    | PLP    | SGD    | DEMON     | PLP    | SGD    |
| 32         | epoch | 34     | 35     | 33     | 30       | 27     | 28     | 33        | 32     | 35     |
|            | Acc/% | 61.23  | 61.27  | 60.65  | 64.49    | 65.35  | 64.28  | 66.06     | 65.85  | 65.06  |
|            | top-1 | 0.3877 | 0.3873 | 0.3935 | 3.551    | 3.465  | 3.572  | 0.3394    | 0.3435 | 0.3494 |
|            | top-5 | 0.1265 | 0.1319 | 0.1328 | 1.184    | 1.078  | 1.073  | 0.1055    | 0.1125 | 0.1157 |
| 64         | epoch | 35     | 35     | 35     | 36       | 37     | 40     | 38        | 35     | 38     |
|            | Acc/% | 59.95  | 59.92  | 59.84  | 63.37    | 64.07  | 63.35  | 62.41     | 62.86  | 61.92  |
|            | top-1 | 0.4005 | 0.4009 | 0.4016 | 3.663    | 3.593  | 3.665  | 0.3759    | 0.3714 | 0.3787 |
|            | top-5 | 0.1383 | 0.1388 | 0.1389 | 1.116    | 1.152  | 1.234  | 0.1296    | 0.1203 | 0.1219 |
| 128        | epoch | 34     | 35     | 37     | 46       | 46     | 48     | 40        | 40     | 44     |
|            | Acc/% | 56.92  | 57.10  | 56.16  | 61.24    | 61.67  | 60.21  | 58.57     | 59.54  | 58.38  |
|            | top-1 | 0.4308 | 0.4290 | 0.4384 | 0.3976   | 0.3833 | 0.3979 | 0.4143    | 0.4046 | 0.4162 |
|            | top-5 | 0.1602 | 0.1600 | 0.1657 | 0.1326   | 0.1326 | 0.1316 | 0.1518    | 0.1416 | 0.1554 |
| 256        | epoch | 37     | 35     | 39     | 62       | 59     | 65     | 52        | 55     | 54     |
|            | Acc/% | 53.43  | 53.47  | 53.13  | 59.75    | 58.93  | 57.84  | 53.40     | 54.58  | 53.65  |
|            | top-1 | 0.4657 | 0.4653 | 0.4687 | 4.035    | 4.194  | 4.157  | 0.4664    | 0.4542 | 0.4635 |
|            | top-5 | 0.1850 | 0.1860 | 0.1875 | 1.580    | 1.607  | 1.623  | 0.1942    | 0.1784 | 0.1845 |

In light of the performance and sensitivity evaluation and analyzation, it can be concluded that the PLP method has good performance and relatively low sensitivity to the change of hyperparameters in most cases, verified the effectiveness of proposed PLP method in optimizing DNN training performance and efficiency.

# 5. Conclusion

In this paper, we proposed a new DNN training method called Parameters Linear Prediction (PLP) method to improve model training performance from a different perspective than mainstream approaches. Specifically, instead of introducing new strategies for momentum or learning rate tuning to optimize training process, the proposed method focuses on the parameters themselves, trying to predict their values in real-time based on their change regularity during training, so as to accelerate model convergence, and improve the generalizability of the model through the introduction of noise injection caused by linear prediction error. It's relatively simple and interpretable.

The proposed PLP method has been transplanted to several commonly used and representative backbones, i.e., Vgg16, Resnet18 and GoogLeNet, and experimented on CIFAR-100 dataset. Results show that it outperforms SGD and is comparable to, or even better than SOTA non-adaptive parameter optimization method. Additionally, the PLP method demonstrates robustness to changes in hyperparameters. Furthermore, the proposed PLP method comprehensively considers issues of real-time needs, data amount and computational complexity, and it's also easy to implement, hardware friendly and no extra hyperparameters are introduced.

However, an undeniable fact is that the proposed method does not always benefit the training process. It can be seen from section 4 that in few particular cases the PLP method performed worse, it is because when there is a mutation in parameters gradient, three-point based linear prediction can't follow the mutation well, which leads to the introduction of excessive errors and then prolongs the training process. Therefore, the improvement of parameter prediction algorithm's ability to handle potential mutations in proposed method is still a problem that needs to be studied.

# Refference

# Figure legends

1. Figure 1: Example of parameters changing tendency during Vgg16 training.

2. Figure 2: Overview of Linear Prediction method. Note that wn_ 1/2/3 and bn_ 1/2/3 represents the n th layer of the model, and 1/2/3 means the parameters obtained from the 1st/2nd/3rd iteration in this cycle. In step b and c $i \in [1, n], i \subseteq N+$

3. Figure 3: Accuracy and comparison of proposed PLP method, DEMON and SGD. 'PLP' refers to networks with proposed Parameters Linear Prediction method, so do the DEMON and SGD.

4. Figure 4: Loss and comparison on validation set of proposed PLP method, DEMON and SGD. The legend "XXX-XXX" in "Loss Diff" refers to the curve shown in figure is the value of "loss of XXX method subtract XXX's loss". The black horizontal solid line is the line of zero, and the red horizontal dashed line represents the average value of the loss difference.

# Tables

1. Table.1 Vgg16 based comparison (factor of loss: 1e-2). Red indicates Top-1 performance.

2. Table.2 Resnet18 based comparison (factor of loss: 1e-2). Red indicates Top-1 performance.

3. Table.3 GoogLeNet based comparison (factor of loss: 1e-2). Red indicates Top-1 performance.

4. Table 4. Average accuracy and top-1/top-5 error results on test set. Red indicates Top-1 performance.

5. Table 5. Performance comparison with different learning rate. Red indicates Top-1 performance. The batch size for these tests is set to128.

6. Table 6. Performance comparison with different batch size. Red indicates Top-1 performance. The learning rate for these tests is set to 0.001.